\documentclass[runningheads]{llncs}
\usepackage[T1]{fontenc}
\usepackage{graphicx}
\usepackage{amssymb}
\usepackage{bm}
\usepackage{mathtools}
\usepackage{multirow}

\begin{document}
\title{Fast Unsupervised Brain Anomaly Detection and Segmentation with Diffusion Models}

\author{Walter H. L. Pinaya \inst{1} \and
Mark S. Graham \inst{1} \and
Robert Gray \inst{2} \and
Pedro F Da Costa \inst{3,4} \and
Petru-Daniel Tudosiu \inst{1} \and
Paul Wright \inst{1} \and
Yee H. Mah \inst{1,5} \and
Andrew D. MacKinnon \inst{6,7} \and
James T. Teo \inst{3,5} \and
Rolf Jager \inst{2} \and
David Werring \inst{8} \and
Geraint Rees \inst{9} \and
Parashkev Nachev \inst{2} \and
Sebastien Ourselin \inst{1} \and
M. Jorge Cardoso \inst{1}}
\authorrunning{Pinaya et al.}
\institute{Department of Biomedical Engineering, School of Biomedical Engineering \& Imaging Sciences, King's College London, UK \and
Institute of Neurology, University College London, UK \and
Institute of Psychiatry, Psychology \& Neuroscience, King's College London, UK \and
Centre for Brain and Cognitive Development, Birkbeck College, UK \and
King’s College Hospital NHS Foundation Trust, UK \and
St George's University Hospitals NHS Foundation Trust, UK \and
Atkinson Morley Regional Neuroscience Centre, UK \and
Stroke Research Centre, UCL Queen Square Institute of Neurology, UK \and
Institute of Cognitive Neuroscience, University College London, UK}
\maketitle
\begin{abstract}
Deep generative models have emerged as promising tools for detecting arbitrary anomalies in data, dispensing with the necessity for manual labelling. Recently, autoregressive transformers have achieved state-of-the-art performance for anomaly detection in medical imaging. Nonetheless, these models still have some intrinsic weaknesses, such as requiring images to be modelled as 1D sequences, the accumulation of errors during the sampling process, and the significant inference times associated with transformers. Denoising diffusion probabilistic models are a class of non-autoregressive generative models recently shown to produce excellent samples in computer vision (surpassing Generative Adversarial Networks), and to achieve log-likelihoods that are competitive with transformers while having fast inference times. Diffusion models can be applied to the latent representations learnt by autoencoders, making them easily scalable and great candidates for application to high dimensional data, such as medical images. Here, we propose a method based on diffusion models to detect and segment anomalies in brain imaging. By training the models on healthy data and then exploring its diffusion and reverse steps across its Markov chain, we can identify anomalous areas in the latent space and hence identify anomalies in the pixel space. Our diffusion models achieve competitive performance compared with autoregressive approaches across a series of experiments with 2D CT and MRI data involving synthetic and real pathological lesions with much reduced inference times, making their usage clinically viable.

\keywords{Denoising Diffusion Probabilistic Models  \and Unsupervised Anomaly Detection \and Out-of-Distribution Detection \and Lesion segmentation \and Neuroimaging}
\end{abstract}
\section{Introduction}
The segmentation of lesions in neuroimaging is an important problem whose solution is of potential value across many clinical tasks, including diagnosis, prognosis, and treatment selection. Ordinarily, segmentation is performed by hand, making this process time-consuming and dependent on human expertise. The development of accurate automatic segmentation methods is therefore crucial to allow the widespread use of precise measurements in clinical routine \cite{porz2014multi,yuh2012quantitative}. Over the last few years, deep generative models have emerged as promising tools for detecting arbitrary lesions and anomalies in data, dispensing with the necessity for either expensive labels or images with anomalies in the training set \cite{baur2021autoencoders,you2019unsupervised,pawlowski2018unsupervised,pinaya2021unsupervised}. These generative models learn the probability density function of normal data and then highlight pathological features as deviations from normality.

Autoregressive models have recently achieved state-of-the-art results in generative modelling \cite{esser2021taming,ramesh2021zero}, and are being used to detect anomalies without supervision in real-world industrial image \cite{wang2020image} and medical imaging \cite{graham2021transformer,pinaya2021unsupervised,patel2021cross}. By factorising the joint distribution of pixel/voxel intensities of an image as a product of conditional distributions $p(\mathbf{x}) = \prod_{i=1}^{n}p(x_i|x_{<i})$ (i.e., in an autoregressive way), the likelihood of images becomes tractable. We can thus directly maximise the expected log-likelihood of the training data, in contrast with Generative Adversarial Networks and Variational Autoencoders. In particular, transformers \cite{vaswani2017attention}, with their attention mechanisms and proven expressivity, have set the state-of-the-art in autoregressive modelling for computer vision \cite{roy2021efficient,jun2020distribution} and in unsupervised anomaly detection for medical imaging \cite{pinaya2021unsupervised}.

Despite their success, transformers still have some weaknesses intrinsic to their autoregressive nature. Due to the unidirectional bias of autoregressive models, the fixed order of sequence elements imposes a perceptually unnatural bias to attention in brain images, constrained to information from preceding elements in the sequence, \cite{pinaya2021unsupervised} employed an ensemble of models with differently ordered versions of a unidimensional input derived from the multidimensional latent representation of vector quantized variational autoencoder (VQ-VAE). Summing across the ensemble improves performance, but at the cost of inference time (the authors used eight transformers to process each 2D image), hindering application in time-critical scenarios. This problem is accentuated with increased data dimensionality (e.g., when analysing 3D data), where even more transformers might be required to achieve good coverage of the image context. In many clinical contexts, such as live data quality control and clinical alerting systems, transformer-based inference times are too slow (>5 min) to make them clinically useful. 

Another issue is the accumulation of prediction errors. The sequential sampling strategy introduces a gap between training and inference, as training relies on so-called teacher-forcing \cite{esser2021imagebart} or exposure bias \cite{schmidt2019generalization}, where the ground truth is provided for each step, whilst inference is performed on previously sampled elements. In anomaly segmentation, this training-inference gap can introduce significant accumulations of errors during the computation of the likelihoods, or in the sampling process involved when “healing” anomalous sequence elements, possibly affecting the quality and coherence of the generated anomaly-corrected images.

In this study, we use denoising diffusion probabilistic models (DDPM or diffusion models for brevity) \cite{ho2020denoising,sohl2015deep}  to create a fast approach that is clinically viable, to eliminate the unidirectional bias, and to avoid accumulated prediction errors during the “healing” process (i.e., the process that remove anomalies from the input image). In essence, DDPMs are trained to iteratively denoise the input by reversing a forward diffusion process that gradually corrupts it. These models are well-founded in principled probabilistic modelling and are able to generate diverse and high-quality images in computer vision \cite{dhariwal2021diffusion,ho2020denoising,nichol2021improved}. Based on recent advances in generative modelling \cite{esser2021imagebart,gu2021vector,rombach2021high}, we use a VQ-VAE \cite{van2017neural} to compress the input image and model its latent space using at diffusion model. This approach uses DDPMs flexibly, rapidly, and efficiently in high dimensional data, such as 3D neuroimaging. In summary, we propose mechanisms to explore the capacities of DDPMs and perform extensive experiments on brain data with synthetic and real lesions.

\section{Background}

\begin{figure}[b!]
\centering
\includegraphics[width=0.7\textwidth]{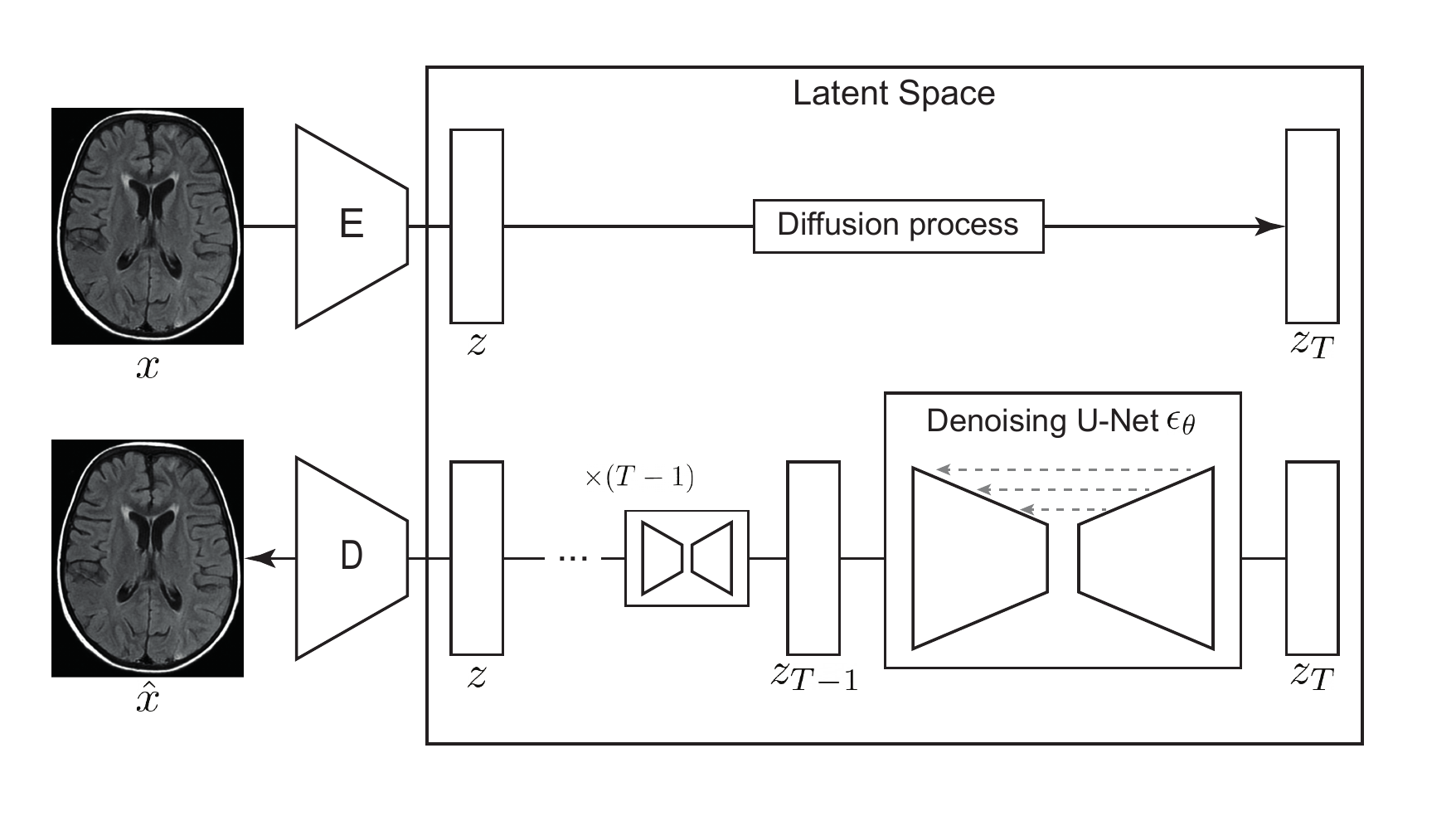}
\caption{The diffusion and reverse processes involved in our anomaly segmentation method, combining a compression model (autoencoder) and a DDPM.} \label{fig1}
\end{figure}

\subsection{Compression model}
Based on previous works \cite{pinaya2021unsupervised,rombach2021high}, we used a VQ-VAE to learn a compact latent representation that offers significantly reduced computational complexity for the diffusion model. The encoder maps any given image, $\mathbf{x}\in\mathbb{R}^{H\times W}$, to a latent representation $E(\mathbf{x})=\mathbf{z}\in\mathbb{R}^{h\times w \times n_z}$. Next, we use the codebook (containing a finite number of embedding vectors $\mathbf{e}_k \in \mathbb{R}^{n_z}, k\in 1, ..., K$, where $K$ is the size of the codebook) to perform an element-wise quantization of each latent variable onto its nearest vector $\mathbf{e}_k$, where $k$ is selected using $k=argmin_j \|E(\mathbf{x})-\mathbf{e}_j\|_2^2$, creating the quantized latent representation $\mathbf{z}$. Then, the decoder $G$ reconstructs the observations $G(\mathbf{z})=\mathbf{\hat{x}}\in\mathbb{R}^{H\times W}$. The encoder $E$, the decoder $G$, and the codebook can be trained end-to-end via $\mathcal{L}_{VQVAE}=\|\mathbf{x}-\mathbf{\hat{x}}\|+\mathcal{L}_{codebook}+\beta\|sg[\mathbf{e}_k]-\mathbf{z}_e\|$ where the operator $sg$ denotes the stop-gradient operation. We used the exponential moving average updates for the codebook loss \cite{van2017neural}. In our experiments, our encoder downsamples the image by a factor of $f$.

\subsection{Denoising Diffusion Probabilistic Models}
In this study, we use the DDPM \cite{ho2020denoising,sohl2015deep} to learn the distribution of the latent representation of healthy brain imaging. The DDPM are latent-variable models consisting of a forward process (or diffusion process - $q(\mathbf{x}_{1:T}|\mathbf{x}_0)$) and a reverse process ($p_{\theta}(\mathbf{x}_{0:T})$) (Fig.~\ref{fig1}). Given a sample from the data distribution $\mathbf{x}_0 \sim q(\mathbf{x}_0)$, the diffusion process gradually destroys the structure of the data via a fixed Markov chain over $T$ steps. Each step in the forward direction is a Gaussian transition defined by $q(\mathbf{x}_t|\mathbf{x}_{t-1})=\mathcal{N}(\mathbf{x}_t;\sqrt{1-\beta_t}\mathbf{x}_{t-1}, \beta_t\mathbf{I})$ where $\beta_t$ follows a fixed variance schedule $\beta_1,...,\beta_T$. By setting $\alpha_t \coloneqq 1-\beta_t$ and $\bar{\alpha}_t \coloneqq \prod_{s=0}^t\alpha_s$, the diffusion process allows sampling $\mathbf{x}_t$ at an arbitrary time step $t$ in an efficient closed form: $q(\mathbf{x}_t|\mathbf{x}_{0})=\mathcal{N}(\mathbf{x}_t;\sqrt{\alpha_t}\mathbf{x}_{0}, (1-\bar{\alpha}_t)\mathbf{I})$. From which any sample can be expressed in terms of some $\bm{\epsilon}  \sim \mathcal{N}(0,\mathbf{I})$ as: $\mathbf{x}_t(\mathbf{x}_0,\bm{\epsilon})=\sqrt{\bar{\alpha}_t}\mathbf{x}_0+\sqrt{1-\bar{\alpha}_t}\bm{\epsilon}$

The reverse process is modelled as a Markov chain which learns to recover the original data $\mathbf{x}_0$ from the noisy input $\mathbf{x}_T$. Since the magnitude of the noise added at each forward step is configured to be small, we can well approximate the true posterior $q(\mathbf{x}_{t-1} |\mathbf{x}_t)$ to a Gaussian distribution, $p_\theta(\mathbf{x}_{t-1}|\mathbf{x}_t)=\mathcal{N}(\mathbf{x}_{t-1};\bm{\mu}_\theta(\mathbf{x}_t,t), \bm{\Sigma}_\theta(\mathbf{x}_t,t))$. Similar to \cite{ho2020denoising}, we use a fixed $\bm{\Sigma}_\theta(\mathbf{x}_t,t)$ obtained according to our variance schedule, and we parametrize our model to predict the cumulative noise $\bm{\epsilon}_0$ that is added to the current intermediate image $\mathbf{x}_t$ to derive $\bm{\mu}_\theta(\mathbf{x}_t,t)=(1/\sqrt{\alpha_t})(\mathbf{x}_t-(\beta_t/\sqrt{\alpha_t-1}) \bm{\epsilon}_\theta(\mathbf{x}_t,t))$. Recently, different methods have been proposed to speed up the reverse process (e.g., Denoising Diffusion Implicit Models - DDIM), reducing by $10\times \sim 50 \times$ the number of necessary reverse steps \cite{song2020denoising}.

Although the data likelihood is intractable, the model can be efficiently trained by maximizing the variational lower bound on the log-likelihood. Following \cite{ho2020denoising}, our loss function can be decomposed as

\begin{multline} \label{eq4}
\mathbb{E}[-\log p_\theta(\mathbf{x}_0)]\leq L\coloneqq \mathbb{E}_q [\underbrace{D_{KL}(q(\mathbf{x}_T|\mathbf{x}_0)\|p(\mathbf{x}_T))}_{L_{T}} \\ + \sum_{t>1}\underbrace{D_{KL}(q(\mathbf{x}_{t-1}|\mathbf{x}_t,\mathbf{x}_0)\|p_{\theta} (\mathbf{x}_{t-1}|\mathbf{x}_t))}_{L_{t-1}} - \underbrace{\log p_{\theta}(\mathbf{x}_1|\mathbf{x}_0)}_{L_0}]
\end{multline} 

\noindent where the term $L_{t-1}$ penalises errors in one reverse step and requires a direct comparison between $p_\theta(\mathbf{x}_{t-1}|\mathbf{x}_{t})$ and its corresponding diffusion process posteriors. We obtain a closed-form expression of the objective since $q(\mathbf{x}_{t-1}|\mathbf{x}_t,\mathbf{x}_0)$ is also a Gaussian distribution; this way, all KL Divergences are comparisons between Gaussians. Using the parametrization from \cite{ho2020denoising}, we obtain the simplified training objective $L_{simple}=\mathbb{E}_{t,\mathbf{x}_0,\bm{\epsilon}}[\| \bm{\epsilon}-\bm{\epsilon}_\theta(\sqrt{\bar{\alpha}_t}\mathbf{x}_0+\sqrt{1-\bar{\alpha}_t}\bm{\epsilon},t)\|]$ where $t$ is sampled uniformly between 1 and $T$ and $\bm{\epsilon}_\theta$ is the learned diffusion model.

\section{Proposed anomaly segmentation method}
Our approach explores the generative model that the DDPM learns from a healthy dataset to guide the brain image healing process. In brief, we use the learned variational lower bound across the DDPM’s Markov chain to identify the latent values that were unlikely to occur in the training set. We replace these unlikely values with more probable ones according to the DDPM, and then we use the latent spatial information to filter the residual maps (obtained from the difference between the original image and the healed image).

After training the VQ-VAE and DDPM on normal data, we use VQ-VAE to obtain the latent representation $\mathbf{z}$ of the test images. After that, we use the DDPM’s forward process to obtain the noisy representations $\mathbf{z}_t$ across the Markov chain. For each step, we use the $L_{t-1}$ values from Eq.~\ref{eq4} to verify how close each reverse step is to the expected Gaussian transition $q(\mathbf{x}_{t-1}|\mathbf{x}_t,\mathbf{x}_0)$. We observed that if the input image is from a healthy subject, the reverse process will only remove the added Gaussian noise, resulting in a low KL Divergence in $L_{t-1}$. However, if the image contains an anomaly, the reverse process removes part of the signal of the original anomalous regions. This signal removal does not follow the expected Gaussian transition, resulting in a high $L_{t-1}$ in the anomalous regions. Using a threshold, we can create a binary mask indicating where the anomalies are and use it to guide the “healing” process.

To find an appropriate threshold, we first obtain the values of $L_{t-1}$ for all our images in the healthy validation set. Instead of using the whole Markov chain, we focus on the steps $t$ inside an intermediary range. As reported in \cite{ho2020denoising}, different $t$ values are responsible for modelling different image features, where higher values are associated with large scale features, and the lower ones are responsible for fine details. In our study, we find that the range of $t=[400,600]$ was less noisy than lower values ($t<150$) and were more specific than high values ($t>800$) to highlight anomalies across different experiments and lesion types. With the $L_{400,600}\in\mathbb{R}^{h\times w \times n_z \times 200}$, we compute the mean values inside the $t$ range and across the $n_z$ dimension. This results in a $\mathbf{v}^k \in \mathbb{R}^{h \times w}$ for each one of the validation samples $k$. Finally, we obtained our 2D threshold using the percentiles 97.5 of validation subjects $percentile_{97.5}(\mathbf{v}^{\text{validation set}})=threshold \in \mathbb{R}^{h \times w}$. Similarly, the evaluated image has its $\mathbf{v}$ obtained, and its values were binarized using $m_{i,j}=1\text{ if }v_{i,j}\geq threshold_{i,j}, 0 \text{ otherwise}$.

The next step of our method is the healing process of the original $\mathbf{z}$. The goal is to inpaint the highlighted regions in $\mathbf{m}$ using the rest of the image as context. For that, we initialize our inpainting process with $t=500$ (in our synthetic lesion tests, we observed that this starting point was enough to heal the lesions). Using the reverse process, we removed “noise” from the regions that have anomalies while keeping the rest as the original $\mathbf{z}_0$, i.e., $\mathbf{z}_{t-1}' \sim p_\theta (\mathbf{z}_{t-1} | \mathbf{m} \odot \mathbf{z}_t'+(1-\mathbf{m}) \odot \mathbf{z}_0)$. This way each step denoises the masked regions a little bit but keeps the rest of the image as original. The resulting $\mathbf{z}_0$  of this process is the latent variable with the anomalies corrected. 

Then, we use the VQ-VAE to decode $\mathbf{z}(=\mathbf{z}_0)$ back to the pixel space $\mathbf{\hat{x}}'$ and obtain the pixel-wise residuals $|\mathbf{x}- \mathbf{\hat{x}}'|$. To clean areas that the DDPM did not specify as anomalous, we upsample our latent mask, smooth it using a Gaussian filter, and multiply it with the residuals. Finally, we can identify regions with anomalies of each brain image from the regions on the final residual maps with high values.

\section{Experiments}
\subsection{Anomaly Segmentation and Detection on Synthetic Anomalies}
In this experiment, we used the MedNIST dataset corrupted with sprites (details in Appendix C). We trained our models with “HeadCT” 9,000 images and we evaluated our method on 100 images contaminated with sprites.

\begin{table}[b]
\caption{Performance of DDPM-based method on synthetic dataset. The performance is measured with best achievable DICE-score ($\lceil DICE \rceil$) and area under the precision-recall curve (AUPRC) on the test set.}\label{tab1}
\begin{center}
\begin{tabular}{lcc}
\hline
  \bfseries Method & \bfseries $\lceil DICE \rceil$  & AUPRC \\
  \hline
  \hline
  AE (Spatial) \cite{baur2021autoencoders}                   & 0.165     & 0.093 \\
  VAE (Dense) \cite{baur2021autoencoders}                    & 0.533     & 0.464 \\
  f-AnoGAN \cite{schlegl2019f}                               & 0.492     & 0.432 \\
  Transformer \cite{pinaya2021unsupervised}                  & 0.768     & 0.808 \\
  Ensemble \cite{pinaya2021unsupervised}                     & 0.895     & \bfseries 0.956 \\
  \hline
  DDPM ($f=4$) (a) [Ours]                                     & 0.777     & 0.810 \\
  DDPM ($f=4$) (b) [Ours]                                     & 0.908     & 0.950 \\
  DDPM ($f=4$) (c) [Ours]                                     & \bfseries 0.920     & 0.955 \\
  \hline
\end{tabular}
\end{center}
\end{table}

Table \ref{tab1} shows how each step of our method improves its performance. Step (a) corresponds to applying a series of reverse steps on $\mathbf{z}$ (no masks, neither masked inpainting). A significant improvement is observed when applying the upsampled mask to the residual maps (step b), and finally, we had the best results with our complete approach (step c). Our single model method has a significant higher performance compared to the transformer  while showing a slightly better performance than the ensemble on the $\lceil DICE \rceil$ but a slightly smaller AUPRC. We also evaluate the anomaly detection performance of our models based on the mean KL Divergence of the images across the whole Markov chain. Using the corrupted images as near out-of-distribution class, our method obtained AUCROC=0.827 and AUPRC=0.702. These values are slightly worse than the transformer-based approach (AUCROC=0.921 and AUPRC=0.707).

\subsection{Anomaly Segmentation on MRI data}
In this experiment, we trained our models on a normative dataset of 15,000 participants with the lowest lesion volume in the UK Biobank (UKB)\cite{sudlow2015uk}. We used FLAIR images, and we evaluated our method on small vessel disease (White Matter Hyperintensities Segmentation Challenge (WMH) \cite{kuijf2019standardized}), tumours (Multimodal Brain Tumor Image Segmentation Benchmark (BRATS) \cite{bakas2018identifying}), demyelinating lesions (Multiple Sclerosis dataset from the University Hospital of Ljubljana (MSLUB) \cite{lesjak2018novel}), and white matter hyperintensities (UKB) (details in Appendix C). Table \ref{tab2} shows that our method performs as well as an ensemble of transformers on the same dataset used to train (i.e., UKB). It performs better than the single transformer on all datasets; however, the ensemble generalizes better.

\begin{table}[t]
\caption{Performance on anomaly segmentation using real 2D MRI lesion data. We measured the performance using the theoretically best possible DICE-score ($\lceil DICE \rceil$). We highlight the best performance in bold and the second best with $^\dagger$.}\label{tab2}
\begin{center}
\begin{tabular}{lcccc}
  \hline
   & \bfseries UKB & \bfseries MSLUB & \bfseries BRATS & \bfseries WMH  \\
  \hline
  \hline
  VAE (Dense) \cite{baur2021autoencoders}       & 0.016     & 0.039     & 0.173     & 0.068 \\
  f-AnoGAN \cite{schlegl2019f}                  & 0.060     & 0.034     & 0.243     & 0.048 \\
  Transformer \cite{pinaya2021unsupervised}     & 0.104     & 0.234     & 0.328     & 0.269 \\
  Ensemble \cite{pinaya2021unsupervised}        & \bfseries 0.232     & \bfseries 0.378     & \bfseries 0.537     & \bfseries 0.429 \\
  \hline
  DDPM ($f=4$) (b) [Ours]                         & 0.208$^\dagger$     & 0.204     & 0.469$^\dagger$     & 0.272 \\
  DDPM ($f=4$) (c) [Ours]                         & \bfseries 0.232     & 0.247$^\dagger$     & 0.398     & 0.298$^\dagger$ \\
  \hline
\end{tabular}
\end{center}
\end{table}

\subsection{Inference Time of Anomaly Segmentation on CT data}
In this experiment, we focused on analysing the inference time of our anomaly segmentation methods in a scenario where time consumption is critical for clinical viability: the analysis of intracerebral haemorrhages (ICH). We trained our models on CT axial slices that did not contain ICH from 200 participants from the CROMIS dataset \cite{wilson2018cerebral}. To evaluate, we used 21 participants from CROMIS and the KCH and CHRONIC \cite{mah2020quantifying} datasets (details in Appendix C).

\begin{table}
\caption{Performance on anomaly segmentation using real 2D CT lesion data. We measured the performance using the theoretically best possible DICE-score ($\lceil DICE \rceil$).}\label{tab3}
\begin{center}
\begin{tabular}{lccccc}
  \hline
   & \bfseries CROMIS & \bfseries KCH & \bfseries CHRONIC & \bfseries Time [s] &   \\
  \hline
  \hline
  VAE (Dense) \cite{baur2021autoencoders}       & 0.185     & 0.353     & 0.171     & $<1$ & \multirow{4}{*}{<1min} \\
  f-AnoGAN \cite{schlegl2019f}                  & 0.146     & 0.292     & 0.099     & $<1$ \\
  DDIM ($f=8$) (d) [Ours]                         & \bfseries 0.286     & \bfseries 0.483     & \bfseries 0.285     & $12$ \\
  DDIM ($f=4$) (d) [Ours]                         & 0.220     & 0.469     & 0.210     & $46$ \\
  \hline
  DDPM ($f=8$) (c) [Ours]                         & 0.284     & 0.473     & 0.297     & $81$ & \multirow{3}{*}{1min$\sim$10min} \\
  DDPM ($f=4$) (c) [Ours]                         & 0.215     & 0.471     & 0.221     & $324$ \\
  Transformer \cite{pinaya2021unsupervised}     & 0.205     & 0.395     & 0.253     & $589$ \\
  \hline
  Ensemble \cite{pinaya2021unsupervised}           & 0.241     & 0.435     & 0.268     & $4907$ & \multirow{3}{*}{>1hour} \\
  Transformer ($f=4$) [Ours]                         & 0.356     & 0.482     & 0.116     & $8047$ \\
  Ensemble ($f=4$) [Ours]                            & 0.471     & 0.631     & 0.122     & $>8000$ \\
  \hline
\end{tabular}
\end{center}
\end{table}

In Table \ref{tab3}, we divide our methods according to the inference time to process 100 slices from the KCH dataset (similar to the number occupied by the brain) on a single GPU (NVIDIA Titan RTX). All slices were fitted in a single minibatch for all models. We analysed different downsampling factors, and we added step (d) of our method where we use $L_{400,600}'$ (with only 50 values evenly spaced) and a DDIM to perform the reverse process (using 50 steps instead of 500). Our methods were able to perform in an acceptable time under 1 minute. Using the DDIM sampler allowed us to significantly improve inference time while keeping a similar performance. All our methods were faster than the transformer-based approaches. As the length of the input sequence grows (changing from $f=8$ to $f=4$), the transformers potentially need to make more forward passes to replace the unlikely tokens. This limits their application of transformers in a higher resolution latent space (which would improve the ability to find smaller lesions). On the other hand, the number of forward passes that DDPM performs is constant for different resolutions, making it easier to scale.

\section{Conclusions}
We proposed a method to use DDPMs to perform anomaly detection and segmentation. The model performed competitively compared with transformers on both synthetic and real data, where it showed a better performance in most cases when compared to a single transformer. Our method holds promise in scenarios where the model prediction has time constraints, especially when using DDIMs. As pointed out in recent studies, anomaly detection methods are essential to obtaining robust performance in clinical settings\cite{graham2021transformer}. We believe that our method's faster inference will help bring high-performance anomaly detection to the clinical front line. DDPMs have just recently caught the attention of the machine learning community, rivalling Generative Adversarial Networks in sample quality and autoregressive models in likelihood scores, built upon a solid theoretical foundation, and fitting within several different theoretical frameworks. We believe that DDPMs have the potential to be even further improved, bringing more advances to anomaly detection in a medical image.

\subsubsection{Acknowledgements}
WHLP, MG, RG, PW, SO, PN and MJC are supported by Wellcome [WT213038/Z/18/Z]. PTD is supported by the EPSRC Research Council, part of the EPSRC DTP, grant Ref: [EP/R513064/1]. YM is supported by an MRC Clinical Academic Research Partnership grant [MR/T005351/1]. PC is supported by SAPIENS Marie Curie Slowdowska Actions ITN N. 814302. PN is also supported by the UCLH NIHR Biomedical Research Centre. MJC and SO are also supported by the Wellcome/EPSRC Centre for Medical Engineering (WT203148/Z/16/Z), and by the GSTT NIHR BRC. This research has been conducted using the UK Biobank Resource (Project number: 58292). The models in this work were trained on NVIDIA Cambridge-1, the UK’s largest supercomputer, aimed at accelerating digital biology.

%
%
%
\bibliographystyle{splncs04}
\bibliography{mybibliography}

\begin{thebibliography}{10}
\providecommand{\url}[1]{\texttt{#1}}
\providecommand{\urlprefix}{URL }
\providecommand{\doi}[1]{https://doi.org/#1}

\bibitem{bakas2018identifying}
Bakas, S., Reyes, M., Jakab, A., Bauer, S., Rempfler, M., Crimi, A., Shinohara,
  R.T., Berger, C., Ha, S.M., Rozycki, M., et~al.: Identifying the best machine
  learning algorithms for brain tumor segmentation, progression assessment, and
  overall survival prediction in the brats challenge. arXiv preprint
  arXiv:1811.02629  (2018)

\bibitem{baur2021autoencoders}
Baur, C., Denner, S., Wiestler, B., Navab, N., Albarqouni, S.: Autoencoders for
  unsupervised anomaly segmentation in brain mr images: a comparative study.
  Medical Image Analysis  \textbf{69},  101952 (2021)

\bibitem{dhariwal2021diffusion}
Dhariwal, P., Nichol, A.: Diffusion models beat gans on image synthesis.
  Advances in Neural Information Processing Systems  \textbf{34} (2021)

\bibitem{esser2021imagebart}
Esser, P., Rombach, R., Blattmann, A., Ommer, B.: Imagebart: Bidirectional
  context with multinomial diffusion for autoregressive image synthesis.
  Advances in Neural Information Processing Systems  \textbf{34} (2021)

\bibitem{esser2021taming}
Esser, P., Rombach, R., Ommer, B.: Taming transformers for high-resolution
  image synthesis. In: Proceedings of the IEEE/CVF Conference on Computer
  Vision and Pattern Recognition. pp. 12873--12883 (2021)

\bibitem{graham2021transformer}
Graham, M.S., Tudosiu, P.D., Wright, P., Pinaya, W.H.L., Teo, J., Jean-Marie,
  U., Mah, Y., J{\"a}ger, R.H., Werring, D., Nachev, P., et~al.:
  Transformer-based out-of-distribution detection for clinically safe
  segmentation  (2021)

\bibitem{gu2021vector}
Gu, S., Chen, D., Bao, J., Wen, F., Zhang, B., Chen, D., Yuan, L., Guo, B.:
  Vector quantized diffusion model for text-to-image synthesis. arXiv preprint
  arXiv:2111.14822  (2021)

\bibitem{ho2020denoising}
Ho, J., Jain, A., Abbeel, P.: Denoising diffusion probabilistic models.
  Advances in Neural Information Processing Systems  \textbf{33},  6840--6851
  (2020)

\bibitem{jun2020distribution}
Jun, H., Child, R., Chen, M., Schulman, J., Ramesh, A., Radford, A., Sutskever,
  I.: Distribution augmentation for generative modeling. In: International
  Conference on Machine Learning. pp. 5006--5019. PMLR (2020)

\bibitem{kuijf2019standardized}
Kuijf, H.J., Biesbroek, J.M., De~Bresser, J., Heinen, R., Andermatt, S., Bento,
  M., Berseth, M., Belyaev, M., Cardoso, M.J., Casamitjana, A., et~al.:
  Standardized assessment of automatic segmentation of white matter
  hyperintensities and results of the wmh segmentation challenge. IEEE
  transactions on medical imaging  \textbf{38}(11),  2556--2568 (2019)

\bibitem{lesjak2018novel}
Lesjak, {\v{Z}}., Galimzianova, A., Koren, A., Lukin, M., Pernu{\v{s}}, F.,
  Likar, B., {\v{S}}piclin, {\v{Z}}.: A novel public mr image dataset of
  multiple sclerosis patients with lesion segmentations based on multi-rater
  consensus. Neuroinformatics  \textbf{16}(1),  51--63 (2018)

\bibitem{mah2020quantifying}
Mah, Y.H., Nachev, P., MacKinnon, A.D.: Quantifying the impact of chronic
  ischemic injury on clinical outcomes in acute stroke with machine learning.
  Frontiers in neurology  \textbf{11}, ~15 (2020)

\bibitem{nichol2021improved}
Nichol, A.Q., Dhariwal, P.: Improved denoising diffusion probabilistic models.
  In: International Conference on Machine Learning. pp. 8162--8171. PMLR (2021)

\bibitem{patel2021cross}
Patel, A., Tudosiu, P.D., Pinaya, W.H.L., Cook, G., Goh, V., Ourselin, S.,
  Cardoso, M.J.: Cross attention transformers for unsupervised whole-body pet
  anomaly detection with multi-modal conditioning  (2021)

\bibitem{pawlowski2018unsupervised}
Pawlowski, N., Lee, M.C., Rajchl, M., McDonagh, S., Ferrante, E., Kamnitsas,
  K., Cooke, S., Stevenson, S., Khetani, A., Newman, T., et~al.: Unsupervised
  lesion detection in brain ct using bayesian convolutional autoencoders
  (2018)

\bibitem{pinaya2021unsupervised}
Pinaya, W.H.L., Tudosiu, P.D., Gray, R., Rees, G., Nachev, P., Ourselin, S.,
  Cardoso, M.J.: Unsupervised brain anomaly detection and segmentation with
  transformers. arXiv preprint arXiv:2102.11650  (2021)

\bibitem{porz2014multi}
Porz, N., Bauer, S., Pica, A., Schucht, P., Beck, J., Verma, R.K., Slotboom,
  J., Reyes, M., Wiest, R.: Multi-modal glioblastoma segmentation: man versus
  machine. PloS one  \textbf{9}(5),  e96873 (2014)

\bibitem{ramesh2021zero}
Ramesh, A., Pavlov, M., Goh, G., Gray, S., Voss, C., Radford, A., Chen, M.,
  Sutskever, I.: Zero-shot text-to-image generation. In: International
  Conference on Machine Learning. pp. 8821--8831. PMLR (2021)

\bibitem{rombach2021high}
Rombach, R., Blattmann, A., Lorenz, D., Esser, P., Ommer, B.: High-resolution
  image synthesis with latent diffusion models. arXiv preprint arXiv:2112.10752
   (2021)

\bibitem{roy2021efficient}
Roy, A., Saffar, M., Vaswani, A., Grangier, D.: Efficient content-based sparse
  attention with routing transformers. Transactions of the Association for
  Computational Linguistics  \textbf{9},  53--68 (2021)

\bibitem{schlegl2019f}
Schlegl, T., Seeb{\"o}ck, P., Waldstein, S.M., Langs, G., Schmidt-Erfurth, U.:
  f-anogan: Fast unsupervised anomaly detection with generative adversarial
  networks. Medical image analysis  \textbf{54},  30--44 (2019)

\bibitem{schmidt2019generalization}
Schmidt, F.: Generalization in generation: A closer look at exposure bias.
  arXiv preprint arXiv:1910.00292  (2019)

\bibitem{sohl2015deep}
Sohl-Dickstein, J., Weiss, E., Maheswaranathan, N., Ganguli, S.: Deep
  unsupervised learning using nonequilibrium thermodynamics. In: International
  Conference on Machine Learning. pp. 2256--2265. PMLR (2015)

\bibitem{song2020denoising}
Song, J., Meng, C., Ermon, S.: Denoising diffusion implicit models. arXiv
  preprint arXiv:2010.02502  (2020)

\bibitem{sudlow2015uk}
Sudlow, C., Gallacher, J., Allen, N., Beral, V., Burton, P., Danesh, J.,
  Downey, P., Elliott, P., Green, J., Landray, M., et~al.: Uk biobank: an open
  access resource for identifying the causes of a wide range of complex
  diseases of middle and old age. PLoS medicine  \textbf{12}(3),  e1001779
  (2015)

\bibitem{van2017neural}
Van Den~Oord, A., Vinyals, O., et~al.: Neural discrete representation learning.
  Advances in neural information processing systems  \textbf{30} (2017)

\bibitem{vaswani2017attention}
Vaswani, A., Shazeer, N., Parmar, N., Uszkoreit, J., Jones, L., Gomez, A.N.,
  Kaiser, {\L}., Polosukhin, I.: Attention is all you need. Advances in neural
  information processing systems  \textbf{30} (2017)

\bibitem{wang2020image}
Wang, L., Zhang, D., Guo, J., Han, Y.: Image anomaly detection using normal
  data only by latent space resampling. Applied Sciences  \textbf{10}(23),
  ~8660 (2020)

\bibitem{wilson2018cerebral}
Wilson, D., Ambler, G., Shakeshaft, C., Brown, M.M., Charidimou, A., Salman,
  R.A.S., Lip, G.Y., Cohen, H., Banerjee, G., Houlden, H., et~al.: Cerebral
  microbleeds and intracranial haemorrhage risk in patients anticoagulated for
  atrial fibrillation after acute ischaemic stroke or transient ischaemic
  attack (cromis-2): a multicentre observational cohort study. The Lancet
  Neurology  \textbf{17}(6),  539--547 (2018)

\bibitem{you2019unsupervised}
You, S., Tezcan, K.C., Chen, X., Konukoglu, E.: Unsupervised lesion detection
  via image restoration with a normative prior. In: International Conference on
  Medical Imaging with Deep Learning. pp. 540--556. PMLR (2019)

\bibitem{yuh2012quantitative}
Yuh, E.L., Cooper, S.R., Ferguson, A.R., Manley, G.T.: Quantitative ct improves
  outcome prediction in acute traumatic brain injury. Journal of neurotrauma
  \textbf{29}(5),  735--746 (2012)

\end{thebibliography}
\end{document}